\documentclass{article}
\usepackage{spconf,amsmath, amssymb, graphicx, CJKutf8, booktabs}
\usepackage[hidelinks]{hyperref}

\title{Dictionary-Constrained Grapheme-to-Phoneme for Unsegmented Languages from LLM-Annotated Data}
\name{Rui Hu, Zhenpeng Zhan, Xiaolong Lin\sthanks{Corresponding author: linxiaolong01@baidu.com}}
\address{Baidu Inc., Shenzhen, China}
\begin{document}
\ninept
% \changefontsize[11pt]{9pt}
%
\maketitle
\begin{abstract}
Grapheme-to-phoneme (G2P) conversion turns raw text into its phonemic form and is an essential part of both text-to-speech (TTS) and automatic speech recognition (ASR) systems. It is required to be fast, stable and context-aware. For unsegmented languages such as Japanese, G2P additionally couples word segmentation with highly context-dependent polyphone disambiguation, and the scarcity of accurately annotated data remains a bottleneck. In this paper, we present a context-aware, segmentation-agnostic neural G2P framework that models the joint segmentation-and-reading hypothesis space, scoring paths of a discriminative conditional random field (CRF) over a dictionary-derived word lattice. To tackle data scarcity, we utilize large language models (LLMs) to generate more than 2 million sentences. Experimental results demonstrate that our method substantially outperforms conventional morphological analyzer-based methods and neural sequence models. On the Joyo-Kanji-Yomi benchmark, our method reaches 99.62\% target word reading accuracy, 0.32\% target word phoneme error rate (PER) and 0.14\% sentence PER.

\end{abstract}
\section{Introduction}
\label{sec:introduction}

Grapheme-to-phoneme (G2P) conversion, which maps text into its phonemic form, serves as a fundamental module in text-to-speech (TTS) and automatic speech recognition (ASR) systems\cite{chen2003conditional, deri2016grapheme, koriyama2026benchmarking}. Although recent state-of-the-art (SOTA) TTS models based on large language models (LLMs)\cite{liao2024fish, hu2026qwen3, zhu2026omnivoice} generate speech directly from raw text at scale, G2P remains useful in TTS systems where controllability and robustness are required\cite{du2025cosyvoice}. It also finds application in TTS evaluation and data filtering\cite{song2024touchtts}.

The complexity of G2P varies fundamentally across language families, dictated by their underlying writing systems and segmentation properties. G2P for alphabetic languages with natural word boundaries, such as English, can largely rely on direct dictionary lookup supplemented by contextual disambiguation for heteronyms\cite{g2pE2019}. Chinese is an unsegmented language with pervasive polyphony. Nevertheless, its strict one-character-to-one-syllable nature allows polyphone disambiguation to be naturally framed as character-level sequence labeling\cite{chen2022g2pw} or lattice-enhanced token classification\cite{zhang2021polyphone} without requiring full-sequence path decoding. In contrast, Japanese presents a far more intricate challenge. Beyond being unsegmented, it employs a mixed writing system of logographic Kanji (\begin{CJK}{UTF8}{ipxm}漢字\end{CJK}) and phonographic Kana (\begin{CJK}{UTF8}{ipxm}カナ\end{CJK}), where a single Kanji character often corresponds to a variable number of syllables and carries highly divergent readings across contexts. Furthermore, lexical compounding frequently triggers irregular phonological changes. For instance, \begin{CJK}{UTF8}{ipxm}恋\end{CJK} (love) can be read as \textit{koi} or \textit{ren}, and \begin{CJK}{UTF8}{ipxm}人\end{CJK} (person) as \textit{hito}, \textit{nin}, or \textit{jin}, yet their compound \begin{CJK}{UTF8}{ipxm}恋人\end{CJK} (lover) becomes \textit{koibito}.

Conventional approaches typically rely on dictionary-driven morphological analyzers\cite{kudo2005mecab, den2008proper, openjtalk, takaoka2018sudachi}. Despite being fast and stable, they capture only local context and rely on explicit segmentation, which propagates errors to reading estimation. More recently, neural sequence models have been applied to Japanese G2P by directly translating graphemes into phonemes\cite{kakegawa2021phonetic, shirahata2026cc}. However, without dictionary constraints, these models can occasionally hallucinate, producing unstable readings. Methods in \cite{DBLP:conf/icassp/HidaHKTSK22, kurihara2024enhancing} utilize pretrained masked language models (MLMs) and dictionaries for better polyphone disambiguation, but they still heavily rely on morphological analyzers for segmentation and feature extraction. Moreover, due to data scarcity, they are often trained on limited human annotations or pseudo-labels from conventional analyzers. Recently, LLMs have prompted initial efforts to evaluate their performance and limits on G2P\cite{qharabagh2025llm, koriyama2026benchmarking}, but training G2P models on large-scale LLM-annotated data remains under-explored.

In this work, we propose a dictionary-constrained, context-aware framework that resolves Japanese G2P through a segmentation-agnostic formulation. Specifically, we formulate Japanese G2P as a conditional random field (CRF) over a dictionary-derived word lattice, marginalizing out latent word segmentations to avoid segmentation error propagation. To power path decoding, we design a character-level DeBERTa-based\cite{he2021deberta} node scorer that injects global contextual information into local lattice nodes for context-dependent polyphone disambiguation. To tackle data scarcity, we utilize frontier LLMs to generate over 2M sentences, taking into consideration the linguistic attributes of Japanese. To validate the effectiveness of the proposed method, we test against the Joyo-Kanji-Yomi (readings of regular-use Kanji) benchmark\cite{liu2026sarashina2} and compare it with several baselines. Experimental results demonstrate that our method achieves 99.62\% target word reading accuracy, 0.32\% target word phoneme error rate (PER) and 0.14\% sentence PER, substantially surpassing conventional methods and neural network counterparts.

% In this work, we propose a dictionary-constrained, context-aware G2P model trained on large amounts of LLM-annotated data. We formulate Japanese G2P as a conditional random field (CRF) over the paths of a word lattice constructed from dictionaries and use a DeBERTa-based\cite{he2021deberta} character-level text encoder to resolve context-dependent polyphone disambiguation. To tackle data scarcity, we utilize frontier LLMs to generate over 2M sentences, taking into consideration the linguistic attributes of Japanese. To validate the effectiveness of the proposed method, we test against the Joyo-Kanji-Yomi (readings of regular-use Kanji) benchmark\cite{liu2026sarashina2} and compare it with several baselines. Experimental results demonstrate that our method achieves 99.62\% target word reading accuracy, 0.32\% target word phoneme error rate (PER) and 0.14\% sentence PER, strongly surpassing conventional methods and neural network counterparts.

\section{Method}
\label{sec:method}

\subsection{Problem formulation}

Conventional methods typically combine a morphological analyzer with a dictionary containing rich lexical information, such as word and transition costs, surface and reading forms, part-of-speech (POS) tags, and conjugation attributes. They first construct a word lattice and search for the path with the lowest accumulated cost. However, for G2P, these methods face two main challenges: (1) they rely on morphologically plausible word segmentation before retrieving readings, causing segmentation errors to propagate to reading prediction; and (2) they primarily rely on local context, making them less effective at disambiguating context-dependent polyphones.

\begin{figure}[t]
    \centering
    \includegraphics[width=0.95\linewidth]{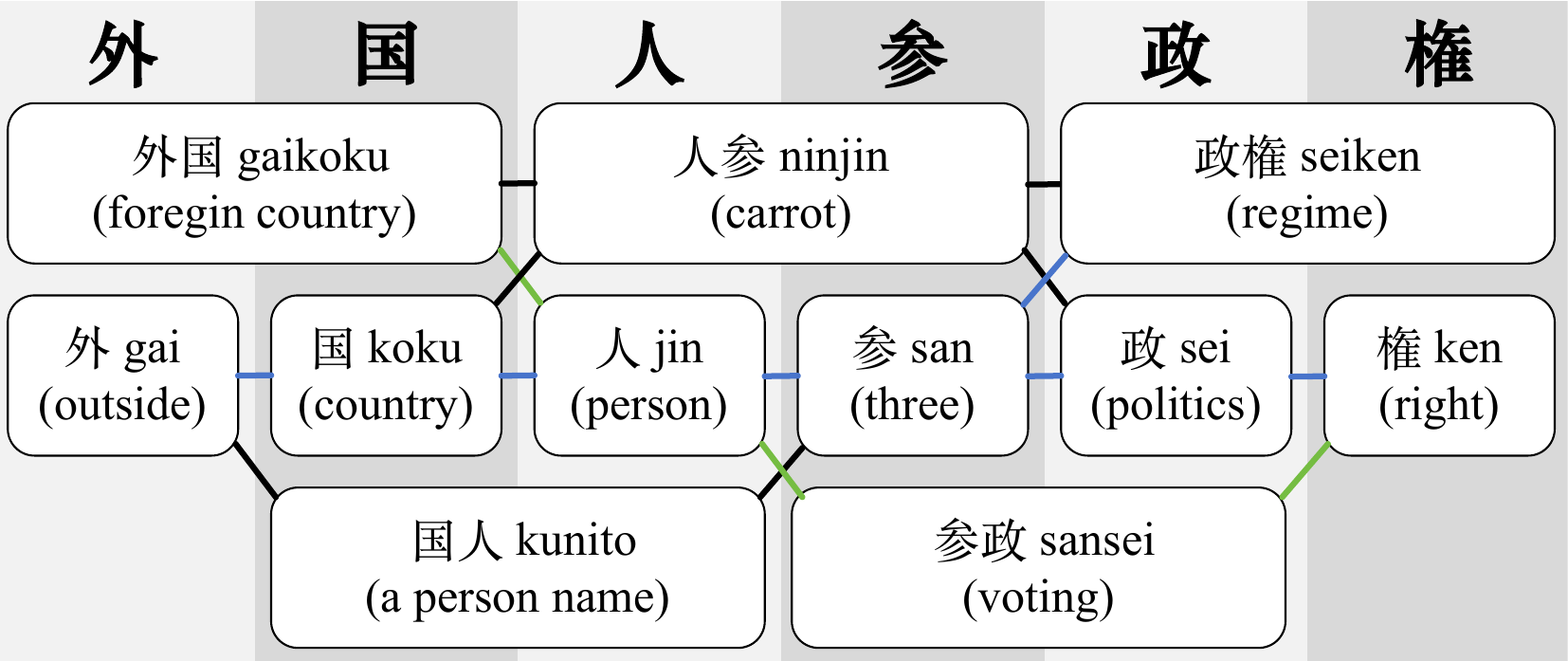}
    \caption{An example of a word lattice (adapted from \cite{morita2015morphological}).}
    \label{fig:lattice}
\end{figure}

Since we only need the full-sentence reading, we relax segmentation and treat any path as valid as long as it yields the correct target reading sequence. In other words, we model the joint segmentation-and-reading hypothesis space supervised by the full-sentence reading. Figure~\ref{fig:lattice} shows an example of a word lattice for the input sentence \begin{CJK}{UTF8}{ipxm}外国人参政権\end{CJK} (voting rights for foreign residents). While morphological analysis strictly favors the path traversing the green edges, for the G2P task, those traversing the blue edges are also considered valid. This relaxation also greatly simplifies data preparation, making LLM annotation more feasible.

\subsection{Model architecture}

To identify nodes that yield correct readings during inference, we design a neural node scorer built on an MLM. As shown in Figure~\ref{fig:model}, the input sentence is first passed through a character-level text encoder initialized with an MLM to obtain contextual representations. Meanwhile, the input sentence is fed into the lattice constructor to build a word lattice. For each node $v$ in the lattice, we slice its corresponding hidden states from the contextual representations based on the position indices of its surface form. The node's reading in Katakana is passed through an embedding layer and a linear projection to obtain reading representations. The two representations are then concatenated along the sequence dimension and fed into several Transformer~\cite{vaswani2017attention} encoder blocks. These concatenated representations then serve as queries in the subsequent Transformer decoder blocks, with the entire contextual representations as keys and values. Note that the self-attention layers in the decoder blocks do not use causal masking. Finally, the fused representations pass through a mean pooling layer and a linear projection to compute a scalar score $s(v)$.

\subsection{Training \& inference}

Let $X$ denote the input sentence and $Y$ denote the target full-sentence reading. During training, we formulate the node scores as a CRF over the lattice DAG. Defining $\mathcal{P}$ as the set of all possible paths in the lattice and $\mathcal{G} \subset \mathcal{P}$ as the subset of valid paths that yield $Y$, the loss is defined as the negative log-marginal-likelihood of the target reading:

\begin{figure}[t]
    \centering
    \includegraphics[width=0.95\linewidth]{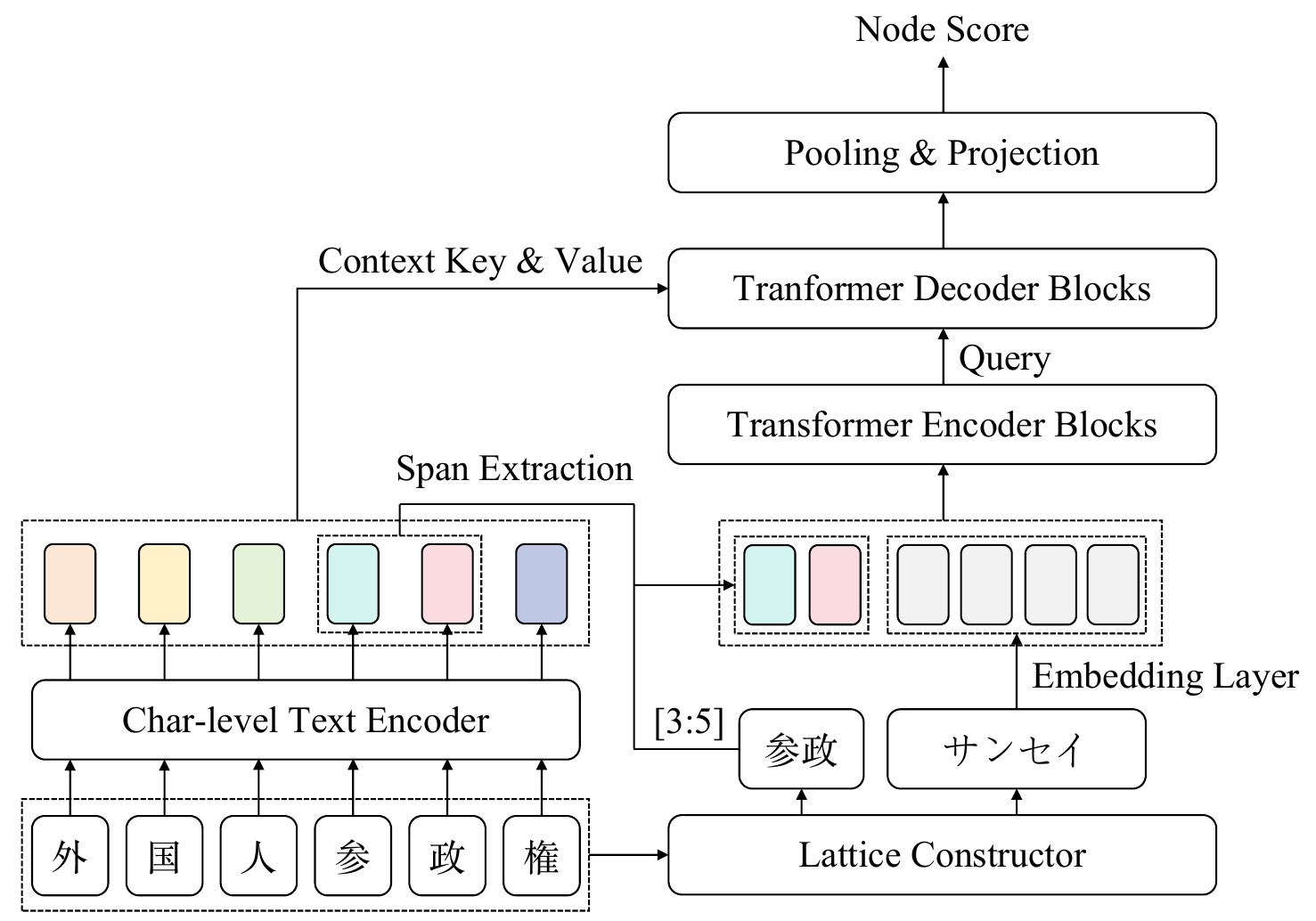}
    \caption{Architecture of the neural node scorer.}
    \label{fig:model}
\end{figure}

\begin{equation}
\mathcal{L}_{\mathrm{crf}}
=
\log \sum_{\pi\in\mathcal{P}} \exp(s(\pi))
-
\log \sum_{\pi\in\mathcal{G}} \exp(s(\pi)),
\label{eq:crf}
\end{equation}

where $s(\pi) = \sum_{v \in \pi} s(v)$ represents the score of a path $\pi$, and both partition functions are computed exactly via a level-wise forward algorithm over the DAG. Because word segmentation is marginalized out in both terms, the objective is indifferent to how a reading is produced and sensitive only to which reading is produced. Notably, unlike conventional CRFs, we omit explicit transition features between adjacent nodes, as the neural node scorer can capture contextual dependencies through the bidirectional text encoder and cross-attention mechanism.

To penalize competing readings over identical spans, we introduce a margin loss over valid-competitor pairs $(v_g, v_c) \in \mathcal{C}$ that share the same character span, where $v_g$ yields a valid reading and $v_c$ represents an incorrect competing reading:
\begin{equation}
\mathcal{L}_{\mathrm{margin}}
=
\frac{1}{|\mathcal{C}|}
\sum_{(v_g, v_c) \in \mathcal{C}}
\max\left(0,\, m - \left(s(v_g) - s(v_c)\right)\right),
\label{eq:margin}
\end{equation}
where $\mathcal{C}$ denotes the set of such competing pairs in the sentence (we define $\mathcal{L}_{\mathrm{margin}} = 0$ if $\mathcal{C} = \emptyset$), and $m > 0$ is a predefined margin hyperparameter. The total objective is then formulated as:
\begin{equation}
\mathcal{L} = \mathcal{L}_{\mathrm{crf}} + \lambda \mathcal{L}_{\mathrm{margin}},
\label{eq:total_loss}
\end{equation}
where $\lambda$ controls the trade-off between the two loss terms. 

During inference, we decode the highest-scoring path via Viterbi search on the lattice DAG and extract its corresponding reading sequence.

\section{Data}
\label{sec:data}

\subsection{Dictionary Preparation}

Two dictionaries are required for lattice construction and LLM annotation, respectively. Both dictionaries are constructed from the widely used UniDic~\cite{den2008proper}, which comprises approximately 870k entries, encompassing common words, proper nouns, and symbols.

For lattice construction, the dictionary should provide the broadest possible coverage, even though there could be a few noisy entries. Therefore, we retain all entries in UniDic. However, we find that many Kanji characters have standalone readings that are not explicitly listed as independent entries in UniDic, but only occur within compound words. For example, the character \begin{CJK}{UTF8}{ipxm}及\end{CJK} has no standalone entry for \textit{kyuu}, despite this reading occurring frequently in compounds such as \begin{CJK}{UTF8}{ipxm}普及\end{CJK} (\textit{fukyuu}) and \begin{CJK}{UTF8}{ipxm}言及\end{CJK} (\textit{genkyuu}). To remedy this, we employ mpaligner~\cite{kubo2011unconstrained} to mine all UniDic entries, yielding approximately 16.6k additional single-character entries.

Conversely, for LLM annotation, the dictionary must be accurate, compact and information-dense. Since UniDic targets fine-grained morphological analysis, words derived from the same base form are fragmented across entries due to conjugation or phonological variations. To streamline this, we group entries by their base form while retaining their POS tags and conjugation attributes. Furthermore, as the primary objective of Japanese G2P is to accurately disambiguate Kanji readings, we filter out entries whose surface forms contain no Kanji characters. Proper nouns are also excluded, as many are context-independent and a single surface form often exhibits multiple equally valid readings. This refinement yields a consolidated dictionary of approximately 151k lexical entries. To improve annotation quality, we partition these into 145k monophonic entries and 6k polyphonic entries.

\subsection{LLM Annotation}

We subsequently leverage Gemini 3.1 Pro,\footnote{\url{https://deepmind.google/models/gemini/pro/}} with thinking level set to high, for downstream data annotation tasks. For each monophonic entry, we prompt the LLM to generate 15 sentences accompanied by their readings, providing the surface form, reading form, POS tags, and conjugation attributes as context. During annotation, the LLM is instructed to first evaluate the validity of the entry and generate instances solely for valid ones, yielding approximately 2.05M sentences.

Since polyphone annotation quality dictates the upper bound of disambiguation performance, we annotate polyphonic entries via a two-stage pipeline: \textit{verification} and \textit{generation}. In the verification stage, the LLM assesses the validity of each candidate reading, clusters valid readings that share nearly identical semantics, provides a concise gloss per cluster, and designates a dominant reading for each, resulting in approximately 10.2k groups. In the generation stage, sentences are synthesized per group by prompting the LLM to construct contexts strictly matching the dominant reading. Besides POS tags and conjugation details, we inject the verification glosses and the competing readings of the same surface form, instructing the LLM to build contexts informative enough for a native speaker to uniquely infer the target reading. We generate 30 sentences per group, producing approximately 306k sentences.

Finally, we apply data filtering for the generated data. First, for each generated sentence and its corresponding reading, we construct a word lattice to verify whether the target reading is reachable. This lattice coverage check filters out roughly 0.76\% of monophonic data and 0.88\% of polyphonic data. Next, in the remaining sentences, we detect any non-dominant or invalid readings identified during polyphone verification: non-dominant readings are mapped to their dominant counterparts, and sentences with invalid readings are discarded. Only 0.06\% of sentences need replacement, and no sentence contains invalid readings, which is reasonable as LLMs rarely generate such readings.

As a result, we obtain a final dataset of approximately 2.04M and 303k sentences from monophonic and polyphonic entries, respectively.

\section{Experiments}
\label{sec:experiments}

\subsection{Evaluation metrics}

We evaluate the proposed method on a revised version of the Joyo-Kanji-Yomi benchmark~\cite{liu2026sarashina2}.\footnote{\url{https://github.com/Parakeet-Inc/Joyo-Kanji-Yomi-Benchmark-Parakeet-Edition}} The benchmark is constructed based on the 2,136 Joyo Kanji (regular-use Kanji) and their 4,512 official readings specified by the Agency for Cultural Affairs of Japan. Each sentence is annotated with its target word's position and reading, alongside the full-sentence reading, totaling 13,536 test sentences. Note that our annotator (Gemini 3.1 Pro) predates the benchmark release, avoiding test-set leakage into the training data.

This revised version rectifies annotation errors in the original benchmark and properly addresses cases where target words admit multiple acceptable readings. Its codebase provides the following evaluation metrics:
\begin{itemize}
    \item \textbf{Accuracy}: The percentage of sentences where the predicted target word reading exactly matches the ground truth.
    \item \textbf{Target PER}: The PER on the target word reading, i.e., the character error rate (CER) over kana sequences, capped at 100\%.
    \item \textbf{Sentence PER}: The PER on the full-sentence reading, defined likewise, capped at 100\%.
\end{itemize}

All metrics are computed per sentence with punctuation removed, then averaged across the benchmark. We report relaxed accuracy and target PER: a prediction matching any acceptable reading is counted as correct.

\subsection{Model details}

The character-level text encoder uses a pretrained Japanese DeBERTa model~\cite{he2021deberta}\footnote{\url{https://huggingface.co/ku-nlp/deberta-v2-base-japanese-char-wwm}} with all parameters unfrozen, followed by a linear projection layer that maps the representations to 512 dimensions. The reading embedding has a dimensionality of 256 and is linearly projected to the 512-dimensional model space. The character-level and reading representations are jointly processed by two Transformer encoder blocks with bidirectional self-attention, followed by two Transformer decoder blocks for cross-attention with the sentence-level context representations. No causal masking is applied in the decoder self-attention. All Transformer blocks use 8 attention heads, a hidden dimension of 512, an FFN dimension of 2048, and a dropout rate of 0.1. During training, the margin $m$ in the margin loss is set to 2.0, and the loss weight $\lambda$ is set to 0.5. We randomly select 5k sentences for validation and use the remaining sentences for training. We train the model for 10 epochs on 4 H100 GPUs. After a warmup period covering 5\% of the total training steps, the peak learning rates for the pretrained DeBERTa and the remaining modules are set to $5\times10^{-5}$ and $3\times10^{-4}$, respectively. We optimize the model with AdamW at a batch size of 640, using a cosine learning rate scheduler with a weight decay of 0.01. The checkpoint with the best validation accuracy is selected for comparison.

\subsection{Baselines}

We compare the proposed method against both conventional morphological analyzers and neural network-based approaches. For the former, we evaluate MeCab~\cite{kudo2005mecab} with UniDic~\cite{den2008proper}, OpenJTalk~\cite{openjtalk}, Juman++~\cite{morita2015morphological}, and Sudachi (Mode C)~\cite{takaoka2018sudachi}. For the latter, we train three neural baselines. The first is an autoregressive model based on ByT5~\cite{xue2022byt5, zhu2022byt5-g2p}, using the \textit{Small} variant backbone comprising 300M parameters. The second is a non-autoregressive CTC-based model~\cite{graves2006connectionist, shirahata2026cc}, whose backbone is initialized from the same Japanese DeBERTa model used as our character-level text encoder. These two are trained via a two-stage pipeline: they are first pretrained on pseudo-labeled readings generated by MeCab with UniDic over 17M sentences from Japanese Wikipedia\footnote{\url{https://huggingface.co/datasets/singletongue/wikipedia-utils}} for 15 epochs, and subsequently fine-tuned on the same LLM-annotated dataset used for our proposed method for 10 epochs. The third is an instruction-tuned LLM, Qwen3-0.6B~\cite{yang2025qwen3}. We directly fine-tune it on the LLM-annotated data for 3 epochs at a learning rate of $5\times10^{-5}$.

Additionally, we include several frontier LLMs, including the Gemini 3 Series\footnote{\url{https://deepmind.google/models/gemini/}} and GPT-5.6 Series,\footnote{\url{https://openai.com/index/gpt-5-6/}} as competitive reference baselines. All LLMs are evaluated under a high reasoning effort setting. To ensure statistical reliability, each LLM is evaluated across 3 independent runs per sample, and we report the averaged performance.

\subsection{Results}

Table~\ref{tab:main} presents the evaluation results on the Joyo-Kanji-Yomi benchmark. Overall, our proposed method substantially outperforms previous approaches and achieves comparable performance with some frontier LLMs, reaching an accuracy of 99.62\%, a target word PER of 0.32\%, and a sentence PER of 0.14\%.

Compared with conventional morphological analyzer-based methods, our model demonstrates a substantial performance leap. This highlights the effectiveness of leveraging contextualized neural scoring over static shallow features for polyphone disambiguation. Take \begin{CJK}{UTF8}{ipxm}米の収穫時期\end{CJK} (the rice harvest season) as an example, Sudachi fails to disambiguate the two readings of the polyphonic character \begin{CJK}{UTF8}{ipxm}米\end{CJK}, \textit{kome} and \textit{bei}, and incorrectly predicts \textit{bei} instead of the more appropriate \textit{kome} in this context.

Our approach also surpasses strong neural baselines, confirming that integrating a constrained word lattice DAG with a neural node scorer provides a stronger structured prior and better mitigates generation instability. For instance, when predicting the reading of \begin{CJK}{UTF8}{ipxm}遠くで稲光が見えた\end{CJK} (I could see a flash of lightning in the distance), the CTC baseline incorrectly reads \begin{CJK}{UTF8}{ipxm}稲光\end{CJK} as \textit{irigabi}, a nonexistent reading, instead of the correct reading \textit{inabikari}. 

Notably, our model achieves competitive accuracy and PER on par with frontier LLMs such as Gemini 3.1 Flash, and approaches the performance of its data generator Gemini 3.1 Pro.

Since our method can only emit readings that are reachable in the lattice, we further report an \textit{oracle} upper bound by selecting the lattice path closest to the ground truth. This yields an accuracy of 100.00\% and a target PER of 0.00\%. These results confirm that lattice coverage is not a bottleneck for Kanji characters. Through case analysis, we find that the remaining sentence-level PER of 0.01\% primarily originates from Arabic numerals, consistent with the limitation discussed in Section~\ref{sec:conclusion}.

\begin{table}[t]
    \centering
    \caption{Results of different G2P methods tested against the Joyo-Kanji-Yomi benchmark. All metrics are reported in \%.}
    \label{tab:main}
    \begin{tabular}{lccc}
    \toprule
    Method & Accuracy & \shortstack{Target PER} & \shortstack{Sentence PER} \\

    \midrule
    Gemini 3.1 Flash & 99.52 & 0.36 & 0.17 \\
    Gemini 3.1 Pro & 99.78 & 0.18 & 0.10 \\
    GPT-5.6-Luna & 99.38 & 0.52 & 0.26 \\
    GPT-5.6-Sol & 99.85 & 0.14 & 0.12 \\

    \midrule
    MeCab +UniDic & 96.66 & 2.82 & 1.23 \\
    OpenJTalk & 96.63 & 3.00 & 0.80 \\
    Sudachi & 97.32 & 2.27 & 1.02 \\
    Juman++ & 90.29 & 9.23 & 2.42 \\

    \midrule
    CTC & 98.09 & 1.22 & 0.42 \\
    ByT5 & 98.45 & 1.34 & 0.38 \\
    Qwen3-0.6B & 98.78 & 1.07 & 0.35 \\

    \midrule
    \textbf{Proposed} & \textbf{99.62} & \textbf{0.32} & \textbf{0.14} \\

    \midrule
    \textit{Oracle} & \textit{100.00} & \textit{0.00} & \textit{0.01} \\
    \bottomrule
    \end{tabular}
\end{table}

\begin{table}[t]
    \centering
    \caption{Results of ablation studies on the Joyo-Kanji-Yomi benchmark. All metrics are reported in \%.}
    \label{tab:ablation}
    \begin{tabular}{lccc}
    \toprule
    Variant & Accuracy & \shortstack{Target PER} & \shortstack{Sentence PER} \\

    \midrule
    W/o margin loss & 99.53 & 0.43 & 0.15 \\

    \midrule
    DeBERTa-tiny & 99.34 & 0.54 & 0.19 \\
    DeBERTa-large & 99.61 & 0.33 & \textbf{0.13} \\

    \midrule
    10\% data & 99.25 & 0.65 & 0.21 \\
    50\% data & 99.52 & 0.43 & 0.16 \\

    \midrule
    Full model & \textbf{99.62} & \textbf{0.32} & 0.14 \\
    
    \bottomrule
    \end{tabular}
\end{table}

We also conduct ablation studies to investigate the impact of the margin loss, model capacity and data scale. The results are summarized in Table~\ref{tab:ablation}. All ablated variants are compared against our full model (\textit{DeBERTa-base} with the margin loss trained on 100\% data). Removing the margin loss leaves the sentence PER nearly unchanged, but causes a noticeable decrease in accuracy and an increase in target PER. This suggests that the margin loss helps the model distinguish the correct candidate node from competing candidate nodes, thereby improving polyphone disambiguation. Varying the size of the pretrained DeBERTa model reveals a clear scaling trend. Reducing the backbone from \textit{DeBERTa-base} (100M) to \textit{DeBERTa-tiny} (10M) degrades all metrics, indicating the importance of sufficient model capacity. In contrast, scaling up to \textit{DeBERTa-large} (300M) yields only a marginal improvement in sentence PER, from 0.14\% to 0.13\%, at the cost of slight degradation in accuracy and target PER. With other settings identical to the full model, training on randomly sampled data subsets (10\% and 50\%) leads to a consistent degradation across all metrics. We conjecture that, under the current full data scale, the \textit{DeBERTa-base} backbone offers a favorable balance between performance and computational efficiency.

\section{Conclusion}
\label{sec:conclusion}

In this paper, we have proposed a context-aware, segmentation-agnostic neural G2P framework for unsegmented languages that scores paths over a dictionary-derived word lattice DAG via a CRF objective, marginalizing out latent segmentations to avoid error propagation. To address data scarcity, we leveraged frontier LLMs to generate over 2M Japanese training sentences. Experimental results demonstrate that our method substantially outperforms conventional morphological analyzers and neural sequence baselines.

Despite these strengths, two limitations point to future work. First, proper nouns were excluded during dictionary filtering to reduce annotation ambiguity and cost, so domain-specific proper nouns require targeted data augmentation. Second, our lattice-based pipeline focuses on Kanji readings and is less suited for Arabic numerals, Latin-script loanwords, units, and mathematical expressions, which are better handled by an upstream text normalization module.

\vfill\pagebreak

\bibliographystyle{IEEEbib}
\bibliography{strings,refs}

\end{document}